\documentclass[10pt,twocolumn,letterpaper]{article}

\usepackage{cvpr}
\usepackage{times}
\usepackage{epsfig}
\usepackage{graphicx}
\usepackage{amsmath}
\usepackage{amssymb}

\usepackage{epstopdf}
\usepackage{enumitem}
\usepackage{multirow}
\newcommand{\calA}{\mathcal{A}}
\newcommand{\calC}{\mathcal{C}}
\newcommand{\calD}{\mathcal{D}}

\usepackage{amsmath,amsfonts}
\usepackage{amssymb,amsopn}
\usepackage{bm} 

\usepackage{amssymb}
\usepackage{amsmath,amsfonts}
\usepackage{amsthm,amsopn}

\usepackage{bm} 

\newcommand{\vct}[1]{\boldsymbol{#1}} 

\newcommand{\ProbOpr}[1]{\mathbb{#1}}

\newcommand{\expect}[2]{%
\ifthenelse{\equal{#2}{}}{\ProbOpr{E}_{#1}}
{\ifthenelse{\equal{#1}{}}{\ProbOpr{E}\left[#2\right]}{\ProbOpr{E}_{#1}\left[#2\right]}}} 
\newcommand{\var}[2]{%
\ifthenelse{\equal{#2}{}}{\ProbOpr{VAR}_{#1}}
{\ifthenelse{\equal{#1}{}}{\ProbOpr{VAR}\left[#2\right]}{\ProbOpr{VAR}_{#1}\left[#2\right]}}} 

\DeclareMathOperator{\argmax}{arg\,max}

\newcommand{\vtheta}{\vct{\theta}}

\newcommand{\vphi}{\vct{\phi}}

\newcommand{\eat}[1]{}

\usepackage[pagebackref=true,breaklinks=true,letterpaper=true,colorlinks,bookmarks=false]{hyperref}

\cvprfinalcopy

\ifcvprfinal\fi
\begin{document}

\title{Learning Answer Embeddings for Visual Question Answering}

\author{
Hexiang Hu$^*$\\
U. of Southern California\\
Los Angeles, CA\\
{\tt\small hexiang.frank.hu@gmail.com}
\and
Wei-Lun Chao\thanks{\hspace{4pt}Equal contributions}\\
U. of Southern California\\
Los Angeles, CA\\
{\tt\small weilunchao760414@gmail.com}
\and
Fei Sha\\
U. of Southern California\\
Los Angeles, CA\\
{\tt\small feisha@usc.edu}
}

\maketitle

\begin{abstract}
We propose a novel probabilistic model for visual question answering (Visual QA). The key idea is to infer two sets of embeddings: one for the image and the question jointly and the other for the answers. The learning objective is to learn the best parameterization of those embeddings such that the correct answer has higher likelihood among all possible answers. In contrast to several existing approaches of treating Visual QA as multi-way classification, the proposed approach takes the semantic relationships (as characterized by the embeddings) among answers into consideration, instead of viewing them as independent ordinal numbers. Thus, the learned embedded function can  be used to embed unseen answers (in the training dataset). These properties make the approach particularly appealing for transfer learning for open-ended Visual QA, where the source dataset on which the model is learned has limited overlapping with the target dataset in the space of answers. We have also developed large-scale optimization techniques for applying the model to datasets with a large number of answers, where the challenge is to properly normalize the proposed probabilistic models. We validate our approach on several Visual QA datasets and investigate its utility for transferring models across datasets. The empirical results have shown that the approach  performs well not only on in-domain learning but also on transfer learning.
\end{abstract}


\section{Introduction}

\begin{figure}[tp]
	\centering
	\vspace{-0.1in}
	\includegraphics[width=0.485\textwidth]{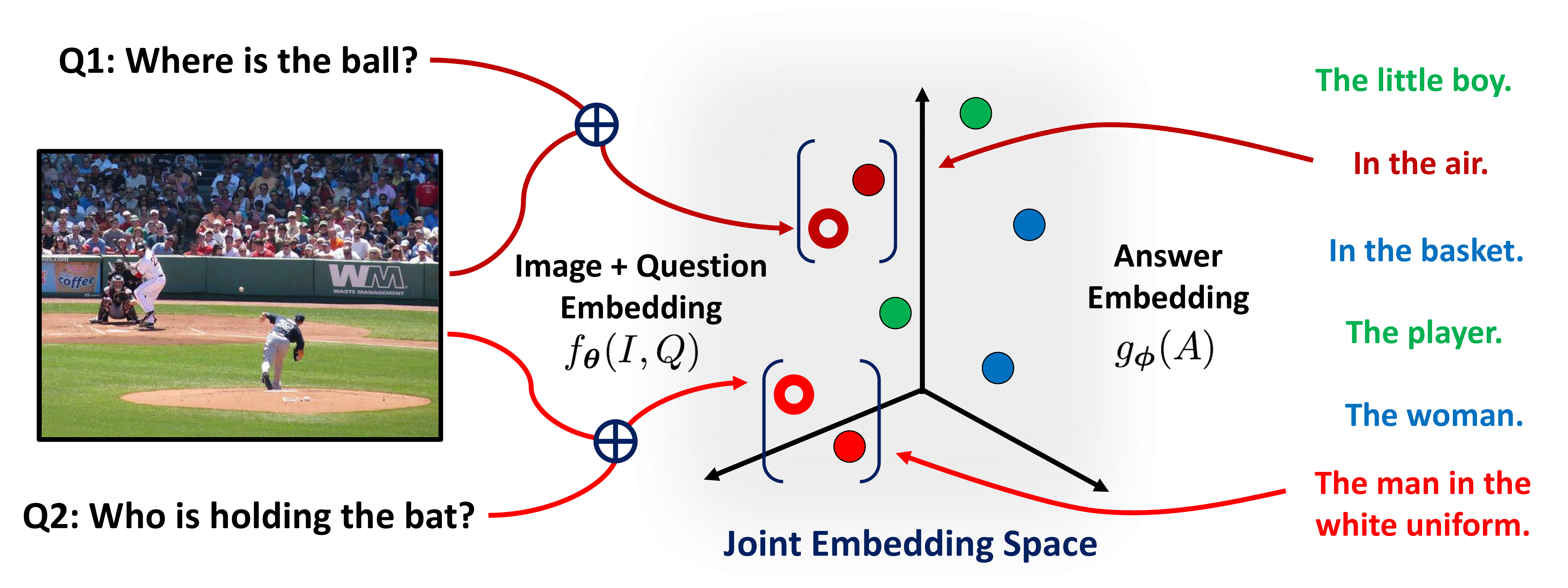}
	\caption{Conceptual diagram of our approach. We learn two embedding functions to transform image question pair $(i, q)$ and (possible) answer $a$ into a joint embedding space. The distance (by inner products) between the embedded $(i, q)$ and $a$ is then measured and the closest $a$ (in red) would be selected as the output answer.}
	\vspace{-10pt}
	\label{fig:illustration}
\end{figure}

Visual question answering (Visual QA) has made significant progress in the last few years. More than 10 datasets have been released for the task~\cite{gupta2017survey, kafle2016visual,wu2016visual}, together with a number of learning models that have been narrowing the gap between the human's performance and the machine's~\cite{yang2016stacked,fukui2016multimodal,lu2016hierarchical,kazemi2017show,xu2016ask}.

In this task, the machine is presented with an image and a related question and needs to output a correct answer. There are several ways of ``outputting'', though. One way is to ask the machine to generate a piece of free-form texts \cite{gao2015you}. However, this often requires humans to decide whether the answer is correct or not. Thus, scaling this type of evaluation to assess a large amount of data (on a large number of models) is challenging.

Automatic evaluation procedures have the advantage of scaling up. There are two major paradigms. One is to use multiple-choice based Visual QA~\cite{zhu2016visual7w,agrawal2016vqa,ren2015exploring}. In this setup, for each pair of image and question, a correct answer is mixed with a set of incorrect answers and the learner optimizes to select the correct one. While popular, it is difficult to design good incorrect answers without bias such that learners are not able to exploit~\cite{chao2017being}. Several recent papers~\cite{chao2017being, jabri2016revisiting} have shown that even when the image or the question is missing, the correct answer can still be identified (using the incidental statistics, \ie, bias, in the data).

The other paradigm that is amenable to automatic evaluation revises the pool of possible answers to be the same for any pair of image and question~\cite{goyal2016making, antol2015vqa}, \ie, open-ended Visual QA. In particular, the pool is composed of most frequent $K$ answers in the training dataset. This has the advantage of framing the task as a multi-way classifier that outputs one of the $K$ categories, with the image and the question as the input to the classifier.

However, while alleviating the bias of introducing incorrect answers that are image and question specific, the open-end Visual QA approaches also suffer from several problems. First, treating the answers as independent categories (as entailed by the multi-way classification) removes the semantic relationship between answers. For example, the answers of ``running'' and ``jogging'' (to the question ``what is the woman in the picture doing?'') are semantically close, so one would naturally infer the corresponding images are visually similar. However, treating ``running'' and ``jogging'' as independent categories ``choice i'' and ``choice j'' would not automatically regularize the learner to ensure the classifier's outputs of visually similar images and semantically similar questions to be semantically close.  In other words, we would desire the outputs of the Visual QA model express semantic proximities aligned with visual and semantic proximities at the inputs. Such alignment will put a strong prior on what the models can learn and prevent them from exploiting biases in the datasets, thus become more robust.

Secondly, Visual QA models learned on one dataset do not transfer to another dataset unless the two datasets share the same space of top $K$ answers---if there is a difference between the two spaces (for example, as ``trivial'' as changing the frequency order of the answers), the classifier will make a substantial number of errors. This is particularly alarming unless we construct a system a prior to map one set of answers to another set, we are likely to have very poor transfer across datasets and would have to train a new Visual QA model whenever we encounter a new dataset. In fact, for two popular Visual QA datasets, about 10\% answers are shared and of top-$K$ answers (where $K<10,000$), only 50\% answers are shared. We refer readers to section~\ref{s_transfer} and Table~\ref{t_transfer_1} for more results.

In this paper, we propose a new learning model to address these challenges. Our main idea is to learn also an embedding of the answers. Together with the (joint embedding) features of image and question in some spaces, the answer embeddings parameterize a probabilistic model describing how the answers are similar to the image and question pair. We learn the embeddings for the answers as well as the images and the questions to maximize the correct answers' likelihood. The learned model thus aligns  the semantic similarity of answers with
the visual/semantic similarity of the image and question pair. Furthermore, the learned model can also embed any unseen answers, thus can generalize from one dataset to another one. Fig. \ref{fig:illustration} illustrates the main idea of our approach.

Our method needs to learn embeddings of hundreds and thousands of answers. Thus to  optimize our probabilistic model, we overcome the challenge by introducing a computationally efficient way of adaptively sampling negative examples in a minibatch.

Our model also has the computational advantage that for each pair of image and question, we only need to compute the joint embedding of image and question for once, irrespective of how many candidate answers one has to examine. On the other end, models such as ~\cite{jabri2016revisiting,fukui2016multimodal} learn a joint embedding of the triplet (image, question and answer) needs to compute embeddings at the linear order of the number of candidate answers. When the number of candidate answers need to be large (to obtain better coverage), such models do not scale up easily.

While our approach is motivated by addressing challenges in open-end Visual QA, the proposed approach trivially includes multiple-choice based Visual QA as a special case and is thus equally applicable. We extensively evaluated our approach on several existing datasets, including Visual7W~\cite{zhu2016visual7w}, VQA2~\cite{goyal2016making}, and Visual Genome~\cite{krishna2016vg}.  We show the gain in performance by our approach over the existing approaches that are based on multi-way classification. We also show the effectiveness of our approach in transferring models trained on one dataset to another. To our best knowledge, we are likely the first to examine the challenging issue of transferability in the open-end Visual QA task\footnote{Our work focuses on the transferability across datasets with different question and answer spaces. We leave visual transferability (e.g., by domain adaptation) as future work.}.

The rest of the paper is organized as follows. Section~\ref{s_notation} introduces the notation and problem setup. Section~\ref{s_main_idea} presents our proposed methods. Section~\ref{s_experiments} shows our empirical results on multiple Visual QA datasets.
\section{Related Work}
\subsection{Visual QA}
In open-end Visual QA, one popular framework of algorithms is to learn a joint image-question embedding and perform multi-way classification (for predicting top-frequency answers) on top~\cite{yu2017multi, anderson2017bottom, ben2017mutan,fukui2016multimodal, yang2016stacked, lu2016hierarchical}. Though such methods naturally limited themselves to answer questions within a fixed (usually small) vocabulary, this framework has been shown to outperform other methods that dedicate for free-form answer generation~\cite{wu2016visual,kafle2016visual}. Different from this line of research, in the multiple-choice setting, algorithms are usually designed to learn a scoring function with the image, question, and answer triplets~\cite{jabri2016revisiting,fukui2016multimodal,shih2016look}. Such methods can take the advantage of answer semantics but fail to scale up inferencing along the increasing number of answer candidates.
Comparing to all previous approaches, our proposed framework leverages the advantages of both worlds, capable of taking the answer semantic into account while remaining efficient. Please refer to section~\ref{s_comparison} for detailed discussion.

\subsection{Learning Aligned Embeddings}
The idea of learning and aligning embeddings has been explored in visual recognition~\cite{frome2013devise, norouzi2014zero}, in which the image and label embeddings are learned. Our work extends it to Visual QA\footnote{We replace the image embedding with a joint one for image-question pairs, and investigate more complicated models for the answer embedding.} for parameterizing and learning a novel probabilistic model. We further propose an efficient optimization technique to handle a large number of candidate answers (e.g., more than 201,000 in Visual Genome~\cite{krishna2016vg}), a situation rarely encountered in visual recognition. 
\section{Methods}
\label{s_method}

In what follows, we describe our approach in detail. We start by describing a general setup for Visual QA and introducing necessary notations. We then introduce the main idea, followed by detailed descriptions of the method and important steps to scale  the method to handling hundreds of thousands negative samples.

\subsection{Setup and Notations}
\label{s_notation}

In the Visual QA task, the machine is given an image $i$ and a question $q$, and is asked to generate an answer $a$. In this work, we focus on the open-ended setting where $a$ is a member of a set $\calA$. This set of candidate answers is intuitively ``the universe of all possible answers''. However, in practice, it is approximated by the top $K$ most frequent correct answers in a training set~\cite{lu2016hierarchical,fukui2016multimodal,yang2016stacked}, plus all the incorrect answers in the dataset (if any).  Another popular setting is multiple-choice based. For each pair of $(i, q)$, the set $\calA$ is different (this set is either automatically generated~\cite{chao2017being} or manually generated~\cite{zhu2016visual7w,agrawal2016vqa}). Without loss of generality, however, we use $\calA$ to represent both. Whenever necessary, we clarify the special handling we would need for $(i, q)$ specific candidate set.

We distinguish two subsets in $\calA$ with respect to a pair $(i, q)$: $\calC$ and $\calD= \calA - \calC$. The set $\calC$ contains all the correct answers for $(i, q)$---it could be a singleton or in some cases, contains multiple \emph{semantically similar} answers to the correct answer (e.g., ``policeman'' to ``police officer''), depending on the datasets. The set $\calD$ contains all the incorrect (or undesired) answers.

A training dataset is thus denoted by a set of $N$ distinctive triplets $D = \{(i_n, q_n, \calC_n)\}$ when only the correct answers are given, or $D = \{(i_n, q_n, \calA_n = \calC_n \cup \calD_n)\}$ when both the correct and incorrect answers are given.

Note that by $i$, $q$ or $a$, we refers to their ``raw'' formats (an image in pixel values, and a question or an answer in its textual forms).

\subsection{Main Idea}
\label{s_main_idea}

Our main idea is motivated by two deficiencies in the current approaches for open-ended Visual QA~\cite{agrawal2016vqa}. In those methods, it is common to construct a $K$-way classifier so that for each $(i, q)$, the classifier outputs $k$ that corresponds to the correct answer (\ie, the $k$-th element in $\calA$ is the correct answer).

However, this classification paradigm cannot capture all the information encoded in the dataset for us to derive better models. First, by equating  two different answers $a_k$ and $a_l$ with the ordinal numbers $k$ and $l$, we lose the semantic kinship between the two. If there are two triplets $(i_m, q_m, a_k \in \calC_m)$ and $(i_n, q_n, a_l \in \calC_n)$ having similar visual appearance between $i_m$ and $i_n$ and similar semantic meaning between $q_m$ and $q_n$, we would expect $a_k$ and $a_l$ to have some degrees of semantic similarity. In a classification framework, such expectation cannot be fulfilled as the assignment of ordinal numbers $k$ or $l$ to either $a_k$ or $a_l$ can be arbitrary such that the difference between $k$ and $l$ does not preserve the similarity between $a_k$ and $a_l$.  However, observing such similarity at both the inputs to the classifier and the outputs of the classifier is beneficial and adds robustness to learning.

The second flaw with the multi-way classification framework is that it does not lend itself to generalize across two datasets with little or no overlapping in the candidate answer sets $\calA$. Unless there is a prior defined mapping between the two sets, the classifier trained on one dataset is not applicable to the other dataset.

We propose a new approach to overcome those deficiencies. The key idea is to learn embeddings of all the data. The embedding functions, when properly parameterized and learned, will preserve similarity and will generalize to unseen answers (in the training data). 

\paragraph{Embeddings} We first define a joint embedding function $f_{\vtheta}(i, q)$ to generate the joint embedding of the pair $i$ and $q$. We also define an embedding function $g_{\vphi}(a)$ to generate the embedding of an answer $a$.  We will postpone to later to explain why we do not learn a function that generates the joint embedding of the triplet.

The embedding functions are parameterized by $\vtheta$ and $\vphi$, respectively. In this work, we use deep learning models such as multi-layer perceptron (MLP) and Stacked Attention Network (SAN)~\cite{yang2016stacked, kazemi2017show} (after removing the classifier at the last layer). In principle, any representation network can be used---our focus is on how to use the embeddings.

\paragraph{Probabilistic Model of Compatibility (PMC)} Given a triplet $(i_n, q_n, a \in \calC_n)$ where $a$ is a correct answer, we define the following probabilistic model
\begin{align}
\label{e_softmax}
p(a |i_n, q_n) =
\cfrac{\exp(f_{\vtheta}(i_n, q_n)^\top g_{\vphi}(a))}{\sum_{a'\in \calA}\exp(f_{\vtheta}(i_n, q_n)^\top g_{\vphi}(a'))}
\end{align}

\paragraph{Discriminative Learning with Weighted Likelihood}
Given the probabilistic model, it is natural to learn the parameters to maximize its likelihood. In our work, we have found the following \emph{weighted} likelihood is more effective
\begin{equation}
\ell = - \sum_n^N \sum_{a \in \calC_n} \sum_{d\in \calA } \alpha(a, d) \log P(d|i_n, q_n),
\label{eWCE}
\end{equation}
where the weighting function $\alpha(a, d)$ measures how much the answer $d$ could contribute to the objective function. A nature design is
\begin{equation}
\alpha (a, d) = \mathbb{I}[a=d],
\end{equation}
where $\mathbb{I}[\cdot]$ is the binary indicator function, taking value of 1 if the condition is true and 0 if false. In this case, the objective function reduces to the standard cross-entropy loss if $\calC_n$ is a singleton.  However, in section~\ref{s_weight_func}, we discuss several different designs.

\subsection{Large-scale Stochastic Optimization}
\label{s_large_scale}

The optimization of eq.~(\ref{eWCE}) is very challenging on real Visual QA datasets. There, the size of $\calA$ can be as large as hundreds of thousands\footnote{In the Visual Genome dataset~\cite{krishna2016vg}, for example, we have more than 201,000 possible answers.}. Thus computing the normalization term of the probability model is a daunting task.

We use a minibatch based stochastic gradient descent procedure to optimize the weighted likelihood.  Specifically, we choose $B$ triplets randomly from $D$ (the training dataset defined in section~\ref{s_notation}) and compute the gradient of the weighted likelihood.

Within a minibatch $(i_b, q_b, \calC_b)$ or $(i_b, q_b, \calC_b\cup \calD_b)$ for $b=1, 2, \cdots B$, we construct a minibatched-universe
\begin{equation}
\calA_B = \bigcup_{b=1}^N (\calC_b \bigcup \calD_b)
\end{equation}
Namely, all the possible answers in the minibatch are used.

However, this ``mini-universe'' might not be a representative sampling of the true ``universe'' $\calA$. Thus, we augment it with \emph{negative sampling}. First we compute the set
\begin{equation}
\bar{\calA}_B = \calA - \calA_B
\end{equation}
and sample $M$ samples from this set. These samples (denoted as $\calA_o$) are mixed with $\calA_B$ to increase the exposure to incorrect answers (\ie negative samples) encountered by the triplets in a minibatch. In short, we use $\calA_0 \bigcup \calA_B$ in lieu of $\calA$ in computing the posterior probability $p(a |i, q)$ and the likelihood.

\subsection{Defining the Weighting Function $\alpha$}
\label{s_weight_func}
We can take advantage of the weighting function $\alpha(a, d)$ to  incorporate external or prior semantic knowledge. For example,  $\alpha(a, d)$ can depend on semantic similiarity scores between $a$ and $d$. Using the WUPS score~\cite{wu1994verbs, malinowski2014multi}, we define the following rule

\begin{equation}
\alpha(a, d)=\left\{
\begin{array}{ll}
1 & \text{ if }\ \textbf{WUPS}(a, d) > \lambda,\\
0  &\text{ otherwise,}
\end{array}\right.
\end{equation}
where $\lambda$ is a  threshold (e.g., 0.9 as in~\cite{malinowski2014multi}). $\alpha(a, d)$ can also be used to scale triplets with a lot of semantic similar answers in $\calC$ (for instance, ``apple'', ''green apple'', ''small apple'' or ``big apple'' are good answers to ``what is on the table?''):
\begin{align}
\alpha(a, d) = \cfrac{\mathbb{I}[a=d]}{|\calC|} \label{e_soft_w}
\end{align}
such that each of these similar answers only contributes to a fraction of the likelihood to the objective function. The idea of eq.~(\ref{e_soft_w}) has been exploited in several recent work~\cite{yu2017beyond,llievski2017simple,kazemi2017show} to boost the performance on VQA~\cite{antol2015vqa} and VQA2~\cite{goyal2016making}.

\subsection{Prediction}

During testing, given the learned $f_{\vtheta}$ and $g_{\vphi}$, for the \textbf{open-ended setting} we can apply the following decision rule
\begin{align}
a^* = \argmax_{a\in \calA} f_{\vtheta}(i, q)^\top g_{\vphi}(a),\label{e_compatible}
\end{align}
to identify the answer to the pair $(i, q)$.

Note that we have the freedom to choose $\calA$ again: it can be the same as the ``universe of answers'' constructed for the training (\ie, the collection of most frequent answers), or a union with all the answers in the validation or testing set. The flexibility is afforded here by using the embedding function $g_{\vphi}$ to embed any texts. Note that in existing open-ended Visual QA, the set $\calA$ is constrained to the most frequent answers, reflecting the limitation of using multi-way classification as a framework for Visual QA tasks.

This decision rule readily extends to the \textbf{multiple-choice setting}, where we just  need to set $\calA$ to include the correct answer and the incorrect answers in each testing triplet.

\subsection{Comparison to Existing Algorithms}
\label{s_comparison}

Most existing Visual QA algorithms (most working on the open-ended setting on VQA~\cite{antol2015vqa} and VQA2~\cite{goyal2016making}) train a multi-way classifier on top of the $f_{\vtheta}$ embedding. The number of classes are set to 1,000 for VQA~\cite{fukui2016multimodal} and around 3,000 for VQA2~\cite{fukui2016multimodal,yu2017beyond,kazemi2017show} of the top-frequency correct answers. These top-frequent answers cover over 90\% of the training and 88\% of the training and validation examples. Those training examples whose correct answers are not in the top-$K$ frequent ones are simply disregarded during training.

There are some algorithms also learning a tri-variable compatibility function $h(i, q, a)$ ~\cite{jabri2016revisiting,fukui2016multimodal,shih2016look}. And the correct answer is inferred by identify $a^*$ such that $h(i, q, a^*)$ is the highest. This type of learning is particularly suitable for multiple-choice based Visual QA. Since the number of candidate answers is small, enumerating all possible $a$ is feasible. However, for open-ended Visual QA tasks, the number of possible answers is very large---computing the function $h()$ for every one of them is costly.

Note that our decision rule relies on computing $f_{\vtheta}(i, q)^\top g_{\vphi}(a)$, a factorized form of the more generic function $h(i, q, a)$. However, precisely due to this factorization, we only need to compute $f_{\vtheta}(i, q)$ just once for every pair $(i, q)$. For $g_{\vphi}(a)$, as long as the model is sufficiently simple, enumerating over many possible $a$ is less demanding than what a generic (and more complex) function $h(i, q, a)$ requires. Indeed, in practice we only need to compute  $g_{\vphi}(a)$ once for any possible $a$\footnote{The answer embedding $g(a)$ for all possible answers (say 100,000) can be pre-computed. At inference we only need to compute the embedding $f(i,q)$ once for an $(i,q)$ pair and perform 100,000 inner products. In contrast, methods like~\cite{jabri2016revisiting,fukui2016multimodal,shih2016look} need to compute $h(i,q,a)$ for 100,000 times. Even if such a function is parameterized with a simple MLP, the computation is much more intensive than an inner product when one has to perform 100,000 times.}. See section~\ref{s_effi} for details.

\section{Experiments}
\label{s_experiments}

We validate our approach on several Visual QA datasets. We start by describing these datasets and the empirical setups. We then report our results. The proposed approach performs very well. It outperforms the corresponding multi-way classification-based approaches where the answers are modeled as independent ordinal numbers. Moreover, it outperforms those approaches in transferring models learned on one dataset to another one.

\subsection{Datasets}

\begin{table}[t]
	\small
	\tabcolsep 2pt
	\caption{Summary statistics of Visual QA datasets.}
	\centering
	\begin{tabular}{c|rrr|rrr|c}\hline
		{Dataset}&  \multicolumn{3}{c|}{\# of Images} &  \multicolumn{3}{c|}{\# of $(i, q, \calC)$ triplets} & {$(|\calC|, |\calD|)$}\\ \cline{2-7}
		{Name} & train & val & test  & train & val & test & per tuple\\
		\hline
		{VQA2}~\cite{goyal2016making} & 83K & 41K & 81K & 443K & 214K & 447K & $(10, 0)$ \\ \hline
		{Visual7W}~\cite{zhu2016visual7w} & 14K & 5K  & 8K & 69K & 28K & 42K & $(1, 3)$ \\ \hline
		{V7W}~\cite{chao2017being} & 14K & 5K  & 8K & 69K & 28K & 42K & $(1, 6)$ \\ \hline
		{qaVG}~\cite{chao2017being}   & 49K & 19K & 29K & 727K & 283K & 433K & $(1, 6)$ \\ \hline
	\end{tabular}
	\label{t_dataset}
	\vskip -10pt
\end{table}

We apply the proposed approach to four datasets. Table~\ref{t_dataset} summarizes their characteristics.
\vspace{-10pt}
\paragraph{VQA2~\cite{goyal2016making}.} The dataset uses images from MSCOCO~\cite{lin2014mscoco} with the same training/validation/testing splits and constructs triplets $(i_n ,q_n, \calC_n)$ of image ($i_n$), question ($q_n$), and correct answers ($\calC_n$) respectively. On average, 6 questions are generated for each image, and each $(i_n, q_n)$ pair is answered by 10 human annotators (i.e, $|\calC_n|=10$). The most frequent one is selected as the single correct answer $t_n$.

\vspace{-10pt}
\paragraph{Visual7W Telling (Visual7W)~\cite{zhu2016visual7w} and V7W~\cite{chao2017being}.} Visual7W uses 47,300 images from MSCOCO \cite{lin2014mscoco} and contains 139,868 $(i_n ,q_n, \calC_n, \calD_n)$ tuples. The set of correct answers $\calC_n$ is a singleton, containing only one answer. Each has 3 incorrect answers generated by humans (i.e., $|\calD_n|=3$). Humans are encouraged to start questions with the 6W words; i.e., ``who'', ``where'', ``how'', ``when'', ``why'', and ``what''.  V7W is a revised version of Visual7W, which has a more carefully designed set of incorrect answers to prevent machines from ignoring the image, or question or both to exploit the bias in the datasets~\cite{chao2017being}. In this dataset, each $(i_n, q_n, \calC_n)$ triplet is associated with 6 auto-generated incorrect answers.

\vspace{-10pt}
\paragraph{Visual Genome (VG)~\cite{krishna2016vg} and qaVG~\cite{chao2017being}.} qaVG~\cite{chao2017being} is a multiple-choice Visual QA dataset derived from VG~\cite{krishna2016vg}. VG contains 101,174 images from MSCOCO \cite{lin2014mscoco} and has 1,445,322 $(i_n, q_n, \calC_n)$ triplets. The set of correct answers $\calC_n$ is a singleton. On average an image is coupled with 14 question-answer pairs. qaVG augments each $(i_n, q_n, \calC_n)$ triplet with 6 auto-generated incorrect answers. The dataset is divided into  50\%, 20\%, and 30\% for training, validation, and testing---each portion is a ``superset'' of the corresponding one in Visual7W or V7W. We train our model on VG~\cite{krishna2016vg} and evaluate it on qaVG~\cite{chao2017being}. 

\begin{table}[t]
	\small
	\centering
	\tabcolsep 4pt
	\caption{The answer coverage of each dataset.}
	\begin{tabular}{c|c|c|c|c}\hline
		& \# of unique answers & \multicolumn{3}{c}{triplets covered by top $K=$ } \\\hline
		{Dataset}  & train/val/test/All & 1,000 & 3,000 & 5,000  \\
		\hline
		{VQA2} & 22K/13K/ - /29K     & 88\% & 93\% & 96\% \\
		{Visual7W}  & 63K/31K/43K/108K  & 57\% & 68\% & 71\% \\
		{VG}   & 119K/57K/79K/201K & 61\% & 72\% & 76\% \\ \hline
	\end{tabular}
	\label{t_vocab_cov}
	\vskip -10pt
\end{table}
\vspace{-10pt}
\paragraph{Answer Coverage within Each Dataset.} In Table~\ref{t_vocab_cov}, We show the number of unique answers in each dataset on each split, together with the portions of question and answer pairs covered by the top-$K$ frequent correct answers from the training set. We observe that the qaVG contains the largest number of answers, followed by Visual7W and VQA2. In terms of coverage, we see that the distribution of answers on VQA2 is the most skewed: over 88\% of training and validation triplets are covered by the top-1000 frequent answers. On the other hand, Visual7W and qaVG needs more than top-5000 frequent answers to achieve a similar coverage.

Thus, a prior, Visual7W and qaVG are ``harder'' datasets, where a multi-way classification-based open-ended Visual QA model will not perform well unless the number of categories is significantly higher (say $\gg 5000$) in order to be able to encounter less frequent answers in the test portion of the dataset---the answers just have a long-tail distribution.
\subsection{Experimental Setup}

\paragraph{Our Model.}
We use two different models to parameterize the embedding function $f_{\vtheta}(i, q)$ in our experiments---Multi-layer Perceptron~\cite{jabri2016revisiting,chao2017being} (MLP) and Stacked Attention Network~\cite{yang2016stacked,kazemi2017show} (SAN). For both models, we first represent each token in the question by the 300-dimensional GloVe vector~\cite{pennington2014glove}, and use the ResNet-152~\cite{he2016deep} to extract the visual features following the exact setting of~\cite{kazemi2017show}. Detailed specifications of each model are as follows.

\begin{itemize}[leftmargin=*]
\item Multi-layer Perceptron (MLP): We represent an image by the 2,048-dimensional vector form the top layer of the ResNet-152 pre-trained on ImageNet~\cite{russakovsky15imagenet}, and a question by the average of the GloVe vectors after a linear transformation followed by tanh non-linearity and dropout. We then concatenate the two features (in total 2,348 dimension), and feed them into a one-layer MLP (4,096 hidden nodes and intermediate dropout), with the output dimensionality of 1,024.

\item Stacked Attention Network (SAN): We represent an image by the $14 \times 14 \times 2048$-dimensional tensor, extracted from the second last layer of the ResNet-152 pre-trained on ImageNet~\cite{russakovsky15imagenet}. See~\cite{lu2016hierarchical} for details. On the other hand, we represent a question by a one layer bidirectional LSTM over GloVe word embeddings. Image and question features are then inputed into the SAN structure for fusion. Specifically, we follow a very similar network architecture presented in~\cite{kazemi2017show}, with the output dimensionality of 1,024.
\end{itemize}

For parameterizing the answering embedding function $g_{\vphi}(a)$, we adopt two architectures:
\textbf{1)} Utilizing a one-layer MLP on \emph{average} GloVe embeddings of answer sequences, with the output dimensionality of 1,024. 
\textbf{2)} Utilizing a two-layer bidirectional LSTM (bi-LSTM) on top of GloVE embeddings of answer sequences.
We use MLP for computing answer embedding by default.
We denote method with bi-LSTM answer embedding with a postfix $\star$ (\eg \textbf{SAN$\star$}). Please refer to our Suppl. Material for more details about architectures and optimization.

In the following, we denote our factorized model applying PMC for optimization as fPMC (cf. eq~(\ref{e_softmax})). We consider variants of fPMC with different architectures (\eg MLP, SAN) for computing $f_{\vtheta}(i, q)$ and $g_{\vphi}(a)$, named as {fPMC(MLP)}, {fPMC(SAN)} and {fPMC(SAN$\star$)}.

\paragraph{Competing Methods.}
We compare our model to multiway classification-based (CLS) models which take either MLP or SAN as  $f_{\vtheta}$. 
We denote them as CLS(MLP) or CLS(SAN). 
We set the number of output classes for CLS model to be top-3,000 frequent training answers for VQA2, and top-5,000 for Visual7W and VG. 
This is a common setup for open-ended Visual QA~\cite{agrawal2016vqa}. 

Meanwhile, we also re-implement approaches that learn a scoring function $h(i, q, a)$ with its input as $(i_n ,q_n, \calC_n)$ triplets~\cite{jabri2016revisiting,chao2017being}.
As such methods are initially designed for multiple-choice datasets, the calibration between positive and negative samples needs to be carefully tuned. 
It is challenging to adapt to `open-end` settings where the number of negative answers scaled up~\footnote{See the Suppl. Material for details}.
Therefore, we adapt them to also utilize our PMC framework for training, which optimize stochastic multi-class cross-entropy with negative answers sampling. 
We name such methods as uPMC (un-factorized PMC) and call its variants as uPMC(MLP) and uPMC(SAN).
We also compare to reported results from other state-of-the-art methods. 

\paragraph{Evaluation Metrics}
The evaluation metric for each dataset is different. For VQA2, the standard metric is to compare the selected answer $a^*$ of a $(i, q)$ pair to the ten corresponding human annotated answers $\calC=\{s_1,\cdots,s_{10}\}$. The performance on such an $(i, q)$ pair is set as follows
\begin{align}
\text{acc}(a^*, \calC) = \max\left\{1, \cfrac{\sum_l \mathbb{I}[a^*=s_l]}{3}\right\}.
\end{align}
We report the average performance over examples in the validation split and test split.

For Visaul7W (or V7W), the performance is measured by the portion of correct answers selected by the Visual QA model from the candidate answer set. The chance for random guess is 25\% (or 14.3\%). 
For VG, we focus on the multiple choice evaluation (on qaVG). We follow the settings proposed by \cite{chao2017being} and measure multiple choice accuracy. The chance for random guess is 14.3\%.
\subsection{Results on Individual Visual QA Datasets}

\begin{table}[t]
	\centering
	\tabcolsep 3pt
	\caption{Results ($\%$) on Visual QA with different settings: open-ended (Top-$K$) and multiple-choice (MC) based for different datasets. The omitted ones are due to their missing in the corresponding work.}
	\label{t_sota}
	\begin{tabular}{c|cc|c|c}\hline
		& {Visual7W} & V7W & {VQA2} & qaVG \\
		{Method} & MC~\cite{zhu2016visual7w} & MC~\cite{chao2017being} & Top-3k~\cite{goyal2016making} &  MC~\cite{chao2017being} \\
		\hline
		{\small{LSTM}~\cite{zhu2016visual7w}}  & 55.6 & - & - & - \\
		{\small{MLP}~\cite{chao2017being}}          & {65.7} & 52.0 & - &  58.5 \\
		{\small{MLP}~\cite{jabri2016revisiting}}     & {67.1} & - & - & - \\
		{\small{C+LSTM}~\cite{goyal2016making}}      & - & - & 54.1 & - \\
		{MCB~\cite{goyal2016making}} & 62.2 & - & 62.3 & -  \\
		{MFB~\cite{yu2017mfb}}       & -  & - & 65.0 & -  \\
		{BUTD~\cite{anderson2017bottom}} & - & - & 65.6 & -  \\
		{MFH~\cite{yu2017beyond}}    & -  & - & {66.8} & - \\
		\hline
	    \multicolumn{5}{c}{Multi-way Classification Based Model (CLS) } \\
		\hline
		{CLS(MLP)}   & 51.6 & 40.9 & 53.5 & 46.9 \\
		{CLS(SAN)}   & 53.7 & 43.6 & 62.4 & 53.0 \\
		\hline
		\multicolumn{5}{c}{Our Probabilistic Model of Compatibility (PMC) } \\
		\hline
		{uPMC(MLP)}   		 & 62.4 & 51.6  & 51.4 & 54.5 \\
		{uPMC(SAN)}  		 & 65.3 & 55.2 & 56.0 & 61.3 \\
		{fPMC(MLP)}    		  & 63.1  & 52.4 & 59.3 & 57.7 \\
		{fPMC(SAN)}    		  & 65.6 & 55.4 & 63.2 & 62.6 \\ \hline
		{fPMC(SAN$\star$)}    & 66.0 & {55.5} & 63.9 & {63.4} \\ \hline
	\end{tabular}
	\vskip -10pt
\end{table}

Table~\ref{t_sota} gives a comprehensive evaluation for most state-of-the-art approaches on four different settings over VQA2 (test-dev), Visual7W, V7W and qaVG\footnote{The omitted ones are due to their missing in the corresponding work. In fact, most existing work only focuses on one or two datasets.}. Among all those settings, our proposed fPMC model outperform the corresponding classification model by a  noticeable margin. Meanwhile, fPMC outperforms uPMC over all settings. Comparing to other state-of-the-art methods, we show competitive performance against most of them. 

In Table~\ref{t_sota}, note that there are differences in the experimental setups in many of the comparison to state-of-the-art methods. For instance,  MLP~\cite{jabri2016revisiting,chao2017being} used either better text embedding or more advanced visual feature, which benefits their result on Visual7W significantly. Under the same configuration, our model has obtained improvement. Besides, most of the state-of-the-art methods on VQA2 fall into the category of classification model that accommodates specific Visual QA settings. They usually explore better architectures for extracting rich visual information~\cite{zhu2016visual7w,anderson2017bottom}, or better fusion mechanisms across multiple modalities~\cite{goyal2016making,yu2017mfb,yu2017beyond}. 
We notice that our proposed PMC model is orthogonal to all those recent advances in multi-modal fusion and neural architectures. More advanced deep learning models can be adapted into our framework as $f_{\vtheta}(i, q)$ (\eg fPMC(MFH)) to achieve superior performance across different settings. This is particularly exemplified by the dominance of SAN over the vanilla MLP model. We leave this for future work.

\subsection{Ablation Studies}

\begin{table}
\centering
\tabcolsep 3.5pt
\caption{The effect of negative sampling ($M=3,000$) on fPMC. The number is the accuracy in each question type on VQA2 (val).}
\label{t_ablation}
\begin{tabular}{cc|cccc}\hline
Method & Mini-Universe & Y/N & Number & Other & All \\ \hline
{MLP} & \multirow{2}{*}{$\calA_B$} & 70.1 & 33.0 & 38.7 & 49.8 \\
{SAN} & & 78.2 & 37.1 & 45.7 & 56.7 \\ \hline
{MLP} & \multirow{2}{*}{$\calA_o \bigcup \calA_B$} & 76.6 & 36.1 & 43.9 & 55.2 \\
{SAN} & & 79.0 & 38.0 & 51.3 & 60.0 \\ \hline
\end{tabular}
\vskip -10pt
\end{table}

\begin{table*}[htb]
	\centering
	\tabcolsep 4pt
	\caption{Results of cross-dataset transfer using either classification-based models or our models (PMC) for Visual QA. ($f_{\vtheta}$ = SAN)}
	\label{t_transfer_2}
	\begin{tabular}{c|cccc|cccc|cccc}\hline
		& \multicolumn{4}{|c}{Visual7W} & \multicolumn{4}{|c}{VQA2} & \multicolumn{4}{|c}{qaVG} \\
		& CLS & uPMC & fPMC & fPMC$\star$ & CLS & uPMC & fPMC & fPMC$\star$ & CLS & fPMC & fPMC & fPMC$\star$ \\
		\hline
		{Visual7W} \
		& 53.7 & 65.3 & 65.6 & \textbf{66.0} {\color{red}$\uparrow$}& 19.1 & 18.5  & \textbf{19.8} {\color{red}$\uparrow$} & 19.1 & 42.8 & 52.2 & \textbf{54.8} {\color{red}$\uparrow$} & 54.3 \\
		{VQA2} \
		& 45.8 & 56.8 & 60.2 &  \textbf{61.7} {\color{red}$\uparrow$} & 59.4 & 56.0  & 60.0 & \textbf{60.9} {\color{red}$\uparrow$} & 37.6 & 51.5  & 54.8 & \textbf{56.8} {\color{red}$\uparrow$} \\
		{qaVG} \
		& 58.9 & 66.0 & 68.4 & \textbf{69.5} {\color{red}$\uparrow$} & 25.6 & 23.6 & 25.8 & \textbf{26.4} {\color{red}$\uparrow$} & 53.0 & 61.2 & 62.6 & \textbf{63.4} {\color{red}$\uparrow$} \\
		\hline
	\end{tabular}
\vskip -10pt
\end{table*}

\paragraph{Importance of Negative Sampling} Our approach is probabilistic, demanding to compute a proper probability over the space of all possible answers. (In contrast, classification-based models limit their output spaces to a pre-determined number, at the risk of not being able to handle unseen answers).

In section~\ref{s_large_scale}, we describe a large-scale optimization technique that allows us to approximate the likelihood by performing negative sampling. Within each mini-batch, we create a mini-universe of all possible answers as the union of all the correct answers (\ie, $\calA_B$). Additionally, we randomly sample $M$ answers from the union of all answers outside of the mini-batch, creating ``an other world'' of all possible answers $\calA_o$. The $\calA_o$ provides richer negative samples to $\calA_B$ and is important to the performance of our model, as shown in Table~\ref{t_ablation}.

\begin{figure}[h]
	\includegraphics[width=0.45\textwidth,trim={0.8cm 0 0.2cm 0},clip]{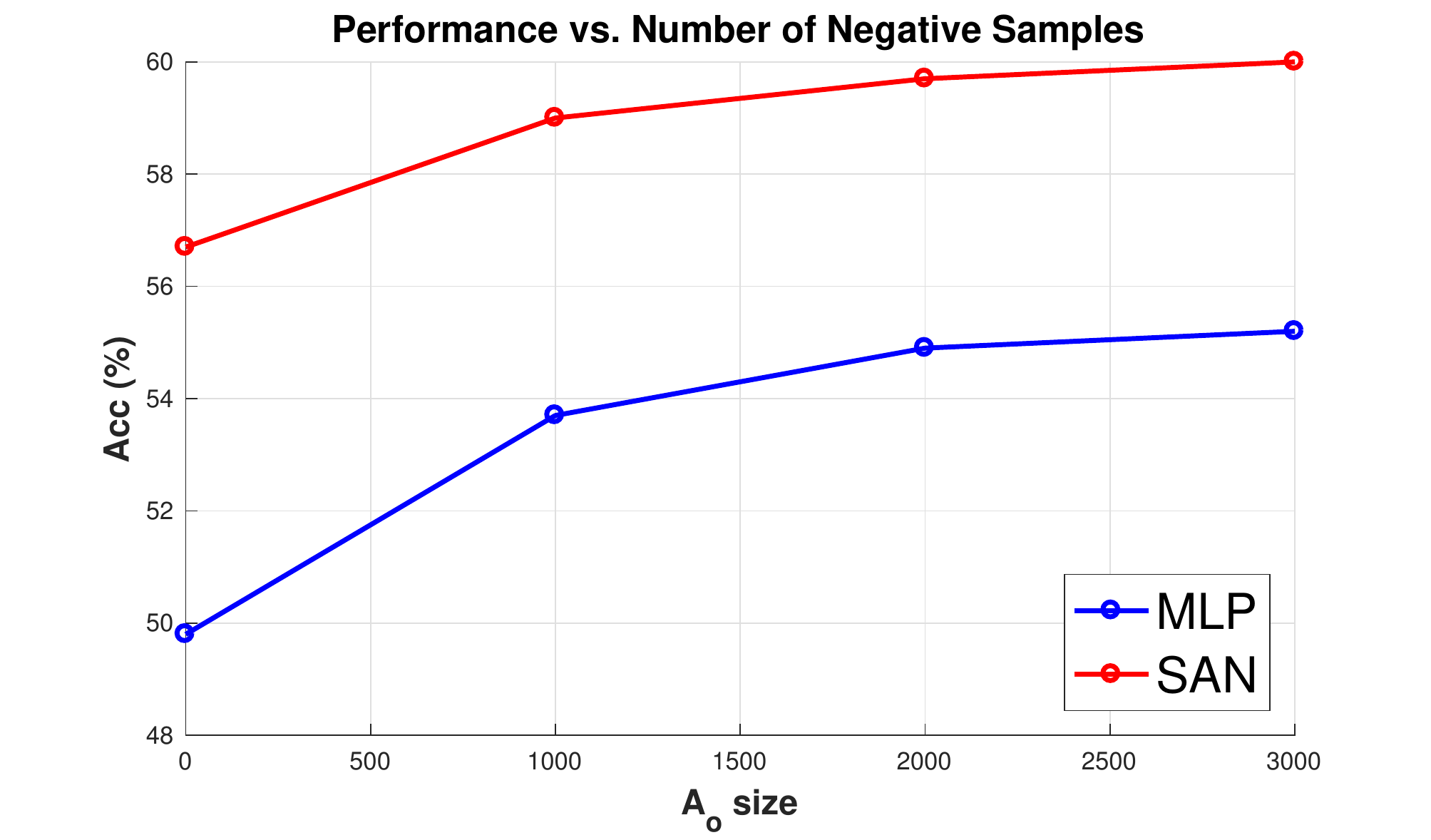}
    \centering
	\caption{Detailed analysis on the size of negative sampling to fPMC(MLP) and fPMC(SAN) at each mini-batch. The reported number is the accuracy on VQA2 (val).}
	\label{f_negative}
	\vskip -10 pt
\end{figure}

We further conducted detailed analysis on the effects of negative sample sizes as shown in Fig.~\ref{f_negative}. With the number of negative samples increasing from 0 to 3,000 for each mini-batch, we observe a increasing trend from the validation accuracy. A significant performance boost is obtained comparing methods with a small number of negative samples to no additional negative samples. The gain then becomes marginal after $\calA_o$ is greater than 2,000.

\paragraph{The Effect of Incorporating Semantic Knowledge in Weighted Likelihood} In section~\ref{s_main_idea}, we have introduced the weighting function $\alpha(a, d)$ to measure how much an incorrect answer $d$ should contribute to the overall objective function. In particular, this weighting function can be used to incorporate prior semantic knowledge about the relationship between a correct answer $a$ and an incorrect answer $d$. We report the details in the Suppl. Material.

\subsection{Transfer Learning Across Datasets}
\label{s_transfer}

\begin{table}[h]
\small
\centering
\tabcolsep 2pt
\caption{The \# of common answers across datasets (training set)}
\begin{tabular}{c|ccccc|c}\hline
& \multicolumn{5}{c|}{Top-$K$ most frequent answers} & Total \# of \\ \cline{2-6}
{Dataset}  & 1K & 3K & 5K & 10K &  all &  unique answers \\
\hline
{VQA2,  Visual7W} & 451 & 1,262 & 2,015 & 3,585 & 10K  & 137K \\
{VQA2,  qaVG} & 495 & 1,328 & 2,057 & 3,643 & 11K  & 149K \\
{Visual7W, qaVG} & 657 & 1,890 & 3,070 & 5,683 & 27K  & 201K \\\hline
\end{tabular}
\label{t_transfer_1}
\vskip-10pt
\end{table}

One important advantage of our method is to be able to cope with unseen answers in the training dataset. This is in stark contrast to multi-way classification based models which will have to skip on those answers as the output categories are selected as top-$K$ most frequent answers from the training dataset.

Thus, classification based models for Visual QA are not amenable to transfer across datasets where there is a large gap between different spaces of answers. Table~\ref{t_transfer_1} illustrates the severity by computing the number of common answers across datasets. On average, about 7\% to 10\% of the unique answers are shared across datasets. If we restrict the number of answers to consider to top 1,000, about 50\% to 65\% answers are shared.  However, top 1000 most frequent answers are in general not enough to cover all the questions in any dataset. Hence, we arrive at the unexciting observation---we can transfer but we can only answer a few questions!

In Table~\ref{t_transfer_2}, we report our results of transferring learned Visual QA model from one dataset (row) to another one (column). For VQA2, we evaluate the open-end accuracy using top-3000 frequent answer candidates on validation set. We evaluate multiple-choice accuracy on the test set of Visual7W and qaVG.

The classification models (CLS) clearly fall behind the performance of our method (uPMC and fPMC)---the red upper arrows signify improvement. In some pairs the improvement is significant (e.g., from 42.8\% to 54.8\% when transferring from Visual7W to qaVG). Furthermore, we noticed that fPMC outperforms uPMC in all transfer settings. 

However, VQA2 seems a particular difficult dataset to be transferred to, from either V7W or qaVG. The improvement from CLS to fPMC is generally small. This is because VQA2 contains a large number of Yes/No answers. For such answers, learning embeddings is not advantageous as there are little semantic meanings to extract from them.

\begin{table}[h]
\centering
\caption{Transferring is improved on the VQA2 dataset without Yes/No answers (and the corresponding questions) ($f_{\vtheta}$ = SAN).}
\begin{tabular}{c|cccc}\hline
Dataset & CLS & uPMC & fPMC & fPMC$\star$\\ \hline
Visual7W  & 31.7 & 29.5 & 33.1 {\color{red}$\uparrow$} & 32.0\\
qaVG & 42.6 & 39.3 & 43.0 & 43.4 {\color{red}$\uparrow$}\\ \hline
\end{tabular}
\label{t_transfer_3}
\end{table}

We perform another study by removing those answers (and associated questions) from VQA2 and report the transfer learning results in Table~\ref{t_transfer_3}. In general, both CLS and fPMC transfer better. Moreover, fPMC improves over CLS by a larger margin than that in Table~\ref{t_transfer_2}. 

To gain a deeper understanding towards which component brings the advantage in transfer learning, we performed additional experiments to analyze the difference on seen/unseen answers. 
At the same time, we include a t-SNE visualization to access the quality of our answer embeddings. We conclude that learned answer embeddings can better capture semantic and syntactic similarities among answers. 
See the Suppl. Material for details on both analysis.

\subsection{Inference Efficiency}
\label{s_effi}
\begin{table}[t]
	\centering
	\caption{Efficiency study among CLS(MLP), uPMC(MLP) and fPMC(MLP). The reported numbers are the average inference time of a mini-batch of 128 ( $|\calC|$ = 1000).}
	\begin{tabular}{c|ccc}\hline
		Method & {CLS(MLP)} & {uPMC(MLP)} & {fPMC(MLP)} \\  \hline
		Time (ms)    & 22.01          & 367.62 		     &  22.14 \\ \hline
	\end{tabular}
	\label{t_efficiency}
	\vskip -10pt
\end{table}

\begin{figure}[h]
	\includegraphics[width=0.455\textwidth]{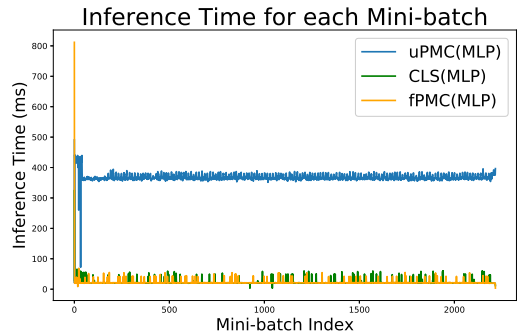}
	\caption{\textbf{Inference time Vs. Mini-batch index.} fPMC(MLP) and CLS(MLP) model are 10x faster than uPMC(MLP) (use PyTorch v0.2.0 + Titan XP + Cuda 8 + Cudnnv5).}
	\label{f_efficiency}
	\vskip-15pt
\end{figure}

Next we study the inference efficiency of the proposed fPMC, uPMC (i.e., triplet based approaches~\cite{jabri2016revisiting,fukui2016multimodal,shih2016look} with PMC) models with the CLS model. For fair comparison, we use the one-hidden-layer MLP model for all approaches, keep $|\calC|=1000$ and mini-batch size to be 128 (uPMC based approach is memory consuming. More candidates require reducing the mini-batch size). We evaluate models on the VQA2 validation set ($\sim$2200 mini-batches) and report the (average) mini-batch inference time. Fig.~\ref{f_efficiency} and Table~\ref{t_efficiency} show that fPMC(MLP) obtains similar performance to CLS(MLP), with at least 10 times faster than uPMC(MLP).

\vspace{-.15in}

\section{Discussion}

We propose a novel approach of learning answer embeddings for the visual question answering (Visual QA) task. The main idea is to learn embedding functions to capture the semantic relationship among answers, instead of treating them as independent categories as in multi-way classification-based models. Besides improving Visual QA results on single datasets, another significant advantage of our approach is to enable better model transfer.  The empirical studies on several datasets have validated our approach. 

Our approach is also ``modular'' in the sense that it can exploit any joint modeling of images and texts (in this case, the questions). An important future direction is to discover stronger multi-modal modeling for this purpose.
  
\footnotesize{ \paragraph{Acknowledgment} This work is partially supported by USC Graduate Fellowship, NSF IIS-1065243, 1451412, 1513966/1632803, 1208500, CCF-1139148, a Google Research Award, an Alfred. P. Sloan Research Fellowship and ARO\# W911NF-12-1-0241 and W911NF-15-1-0484. }

{\small
\bibliographystyle{ieee}
\bibliography{vqa}

\begin{thebibliography}{10}\itemsep=-1pt

\bibitem{agrawal2016vqa}
A.~Agrawal, J.~Lu, S.~Antol, M.~Mitchell, C.~Lawrence~Zitnick, D.~Parikh, and
  D.~Batra.
\newblock Vqa: Visual question answering.
\newblock {\em IJCV}, 2016.

\bibitem{anderson2017bottom}
P.~Anderson, X.~He, C.~Buehler, D.~Teney, M.~Johnson, S.~Gould, and L.~Zhang.
\newblock Bottom-up and top-down attention for image captioning and vqa.
\newblock In {\em CVPR}, 2018.

\bibitem{antol2015vqa}
S.~Antol, A.~Agrawal, J.~Lu, M.~Mitchell, D.~Batra, C.~Lawrence~Zitnick, and
  D.~Parikh.
\newblock Vqa: Visual question answering.
\newblock In {\em ICCV}, 2015.

\bibitem{ben2017mutan}
H.~Ben-younes, R.~Cadene, M.~Cord, and N.~Thome.
\newblock Mutan: Multimodal tucker fusion for visual question answering.
\newblock In {\em ICCV}, 2017.

\bibitem{chao2017being}
W.-L. Chao, H.~Hu, and F.~Sha.
\newblock Being negative but constructively: Lessons learnt from creating
  better visual question answering datasets.
\newblock In {\em NAACL}, 2018.

\bibitem{frome2013devise}
A.~Frome, G.~S. Corrado, J.~Shlens, S.~Bengio, J.~Dean, T.~Mikolov, et~al.
\newblock Devise: A deep visual-semantic embedding model.
\newblock In {\em NIPS}, 2013.

\bibitem{fukui2016multimodal}
A.~Fukui, D.~H. Park, D.~Yang, A.~Rohrbach, T.~Darrell, and M.~Rohrbach.
\newblock Multimodal compact bilinear pooling for visual question answering and
  visual grounding.
\newblock In {\em EMNLP}, 2016.

\bibitem{gao2015you}
H.~Gao, J.~Mao, J.~Zhou, Z.~Huang, L.~Wang, and W.~Xu.
\newblock Are you talking to a machine? dataset and methods for multilingual
  image question.
\newblock In {\em NIPS}, pages 2296--2304, 2015.

\bibitem{goyal2016making}
Y.~Goyal, T.~Khot, D.~Summers-Stay, D.~Batra, and D.~Parikh.
\newblock Making the v in vqa matter: Elevating the role of image understanding
  in visual question answering.
\newblock In {\em CVPR}, 2017.

\bibitem{gupta2017survey}
A.~K. Gupta.
\newblock Survey of visual question answering: Datasets and techniques.
\newblock {\em arXiv preprint arXiv:1705.03865}, 2017.

\bibitem{he2016deep}
K.~He, X.~Zhang, S.~Ren, and J.~Sun.
\newblock Deep residual learning for image recognition.
\newblock In {\em CVPR}, 2016.

\bibitem{llievski2017simple}
I.~{Ilievski} and J.~{Feng}.
\newblock A simple loss function for improving the convergence and accuracy of
  visual question answering models.
\newblock In {\em CVPR Workshop}, 2017.

\bibitem{jabri2016revisiting}
A.~Jabri, A.~Joulin, and L.~van~der Maaten.
\newblock Revisiting visual question answering baselines.
\newblock In {\em ECCV}, 2016.

\bibitem{kafle2016visual}
K.~Kafle and C.~Kanan.
\newblock Visual question answering: Datasets, algorithms, and future
  challenges.
\newblock {\em Computer Vision and Image Understanding}, 163:3--20, 2017.

\bibitem{kazemi2017show}
V.~Kazemi and A.~Elqursh.
\newblock Show, ask, attend, and answer: A strong baseline for visual question
  answering.
\newblock {\em arXiv preprint arXiv:1704.03162}, 2017.

\bibitem{kingma2014adam}
D.~Kingma and J.~Ba.
\newblock Adam: A method for stochastic optimization.
\newblock In {\em ICLR}, 2015.

\bibitem{krishna2016vg}
R.~Krishna, Y.~Zhu, O.~Groth, J.~Johnson, K.~Hata, J.~Kravitz, S.~Chen,
  Y.~Kalantidis, L.-J. Li, D.~A. Shamma, M.~Bernstein, and L.~Fei-Fei.
\newblock Visual genome: Connecting language and vision using crowdsourced
  dense image annotations.
\newblock {\em IJCV}, 2017.

\bibitem{lin2014mscoco}
T.-Y. Lin, M.~Maire, S.~Belongie, J.~Hays, P.~Perona, D.~Ramanan,
  P.~Doll{\'a}r, and C.~L. Zitnick.
\newblock Microsoft coco: Common objects in context.
\newblock In {\em ECCV}, 2014.

\bibitem{lu2016hierarchical}
J.~Lu, J.~Yang, D.~Batra, and D.~Parikh.
\newblock Hierarchical question-image co-attention for visual question
  answering.
\newblock In {\em NIPS}, pages 289--297, 2016.

\bibitem{malinowski2014multi}
M.~Malinowski and M.~Fritz.
\newblock A multi-world approach to question answering about real-world scenes
  based on uncertain input.
\newblock In {\em NIPS}, 2014.

\bibitem{norouzi2014zero}
M.~Norouzi, T.~Mikolov, S.~Bengio, Y.~Singer, J.~Shlens, A.~Frome, G.~S.
  Corrado, and J.~Dean.
\newblock Zero-shot learning by convex combination of semantic embeddings.
\newblock In {\em ICLR}, 2014.

\bibitem{pennington2014glove}
J.~Pennington, R.~Socher, and C.~D. Manning.
\newblock Glove: Global vectors for word representation.
\newblock In {\em EMNLP}, 2014.

\bibitem{ren2015exploring}
M.~Ren, R.~Kiros, and R.~Zemel.
\newblock Exploring models and data for image question answering.
\newblock In {\em NIPS}, 2015.

\bibitem{russakovsky15imagenet}
O.~Russakovsky, J.~Deng, H.~Su, J.~Krause, S.~Satheesh, S.~Ma, Z.~Huang,
  A.~Karpathy, A.~Khosla, M.~Bernstein, A.~C. Berg, and L.~Fei-Fei.
\newblock {ImageNet} large scale visual recognition challenge.
\newblock {\em IJCV}, 115(3):211--252, 2015.

\bibitem{shih2016look}
K.~J. Shih, S.~Singh, and D.~Hoiem.
\newblock Where to look: Focus regions for visual question answering.
\newblock In {\em CVPR}, 2016.

\bibitem{wu2016visual}
Q.~Wu, D.~Teney, P.~Wang, C.~Shen, A.~Dick, and A.~v.~d. Hengel.
\newblock Visual question answering: A survey of methods and datasets.
\newblock {\em Computer Vision and Image Understanding}, 163:21--40, 2017.

\bibitem{wu1994verbs}
Z.~Wu and M.~Palmer.
\newblock Verbs semantics and lexical selection.
\newblock In {\em ACL}, 1994.

\bibitem{xie2017aggregated}
S.~Xie, R.~Girshick, P.~Doll{\'a}r, Z.~Tu, and K.~He.
\newblock Aggregated residual transformations for deep neural networks.
\newblock In {\em CVPR}, 2017.

\bibitem{xu2016ask}
H.~Xu and K.~Saenko.
\newblock Ask, attend and answer: Exploring question-guided spatial attention
  for visual question answering.
\newblock In {\em ECCV}, 2016.

\bibitem{yang2016stacked}
Z.~Yang, X.~He, J.~Gao, L.~Deng, and A.~Smola.
\newblock Stacked attention networks for image question answering.
\newblock In {\em CVPR}, 2016.

\bibitem{yu2017multi}
Z.~Yu, J.~Yu, J.~Fan, and D.~Tao.
\newblock Multi-modal factorized bilinear pooling with co-attention learning
  for visual question answering.
\newblock In {\em ICCV}, 2017.

\bibitem{yu2017beyond}
Y.~Zhou, Y.~Jun, X.~Chenchao, F.~Jianping, and T.~Dacheng.
\newblock Beyond bilinear: Generalized multi-modal factorized high-order
  pooling for visual question answering.
\newblock {\em arXiv preprint arXiv:1708.03619}, 2017.

\bibitem{yu2017mfb}
Y.~Zhou, Y.~Jun, F.~Jianping, and T.~Dacheng.
\newblock Multi-modal factorized bilinear pooling with co-attention learning
  for visual question answering.
\newblock In {\em ICCV}, 2017.

\bibitem{zhu2016visual7w}
Y.~Zhu, O.~Groth, M.~Bernstein, and L.~Fei-Fei.
\newblock Visual7w: Grounded question answering in images.
\newblock In {\em CVPR}, 2016.

\end{thebibliography}
}

\clearpage
\appendix
\begin{center}
	\textbf{\Large Supplementary Material}
\end{center}

In this Supplementary Material, we provide details omitted in the main paper:

\begin{itemize}
  \item Section~\ref{s_details}: Implementation details (Section 4.2 of
  the main paper).
  \item Section~\ref{s_PMC_triplet}: uPMC Vs. Triplet-based Methods (Section 4.2 of the main paper).
  \item Section~\ref{s_semantics}: Effects of incorporating semantic knowledge in weighted likelihood (Section 4.4 of the main paper).
  \item Section~\ref{s_unseen}: Analysis with seen/unseen answers (Section 4.5 of the main paper).
  \item Section~\ref{s_visualize}: Visualization of answer embeddings (Section 4.5 of the main paper).
  \item Section~\ref{s_baseline}: Analysis on answer embeddings.
\end{itemize}

\section{Implementation Details}
\label{s_details}

In this section, we provide more details about the architectures of the stacked attention network (SAN)~\cite{kazemi2017show,yang2016stacked} and the multi-layer perceptron (MLP) used for $f_{\vtheta}(i, q)$ and $g_{\vphi}(a)$ in the main paper (section 4.2). 

\paragraph{ MLP as $f_{\vtheta}(i, q)$ and $g_{\vphi}(a)$}
As mentioned in the main paper, a one-hidden-layer MLP (with the hidden dimension of 4,096 and output dimension of 1,024) is used for both $f_{\vtheta}(i, q)$ and $g_{\vphi}(a)$. The question $q$ or answer $a$ is represented by the average of word embeddings. Concretely, we compute the average of the pre-trained GloVe~\cite{pennington2014glove} vectors of words in question or answer. We then input this question vector (concatenated with the visual feature) or answer vector to the specified MLP, for obtaining output embedding. To enable better generalization on unseen answers and across datasets, we keep the average GloVe word embeddings for answer fixed in all our experiments. For the word embedding on questions, we fine-tune it as this leads to better empirical results. To represent the image feature $i$, we extract the activations from the last convolution layer of a 152-layer ResNet~\cite{he2016deep} pre-trained on ImageNet~\cite{russakovsky15imagenet}, and average them over the spatial extent to obtain a 2,048 dimensional feature vector. 
 
\begin{figure}[t]
	\centering
	\vspace{-0.1in}
	\includegraphics[width=0.475\textwidth]{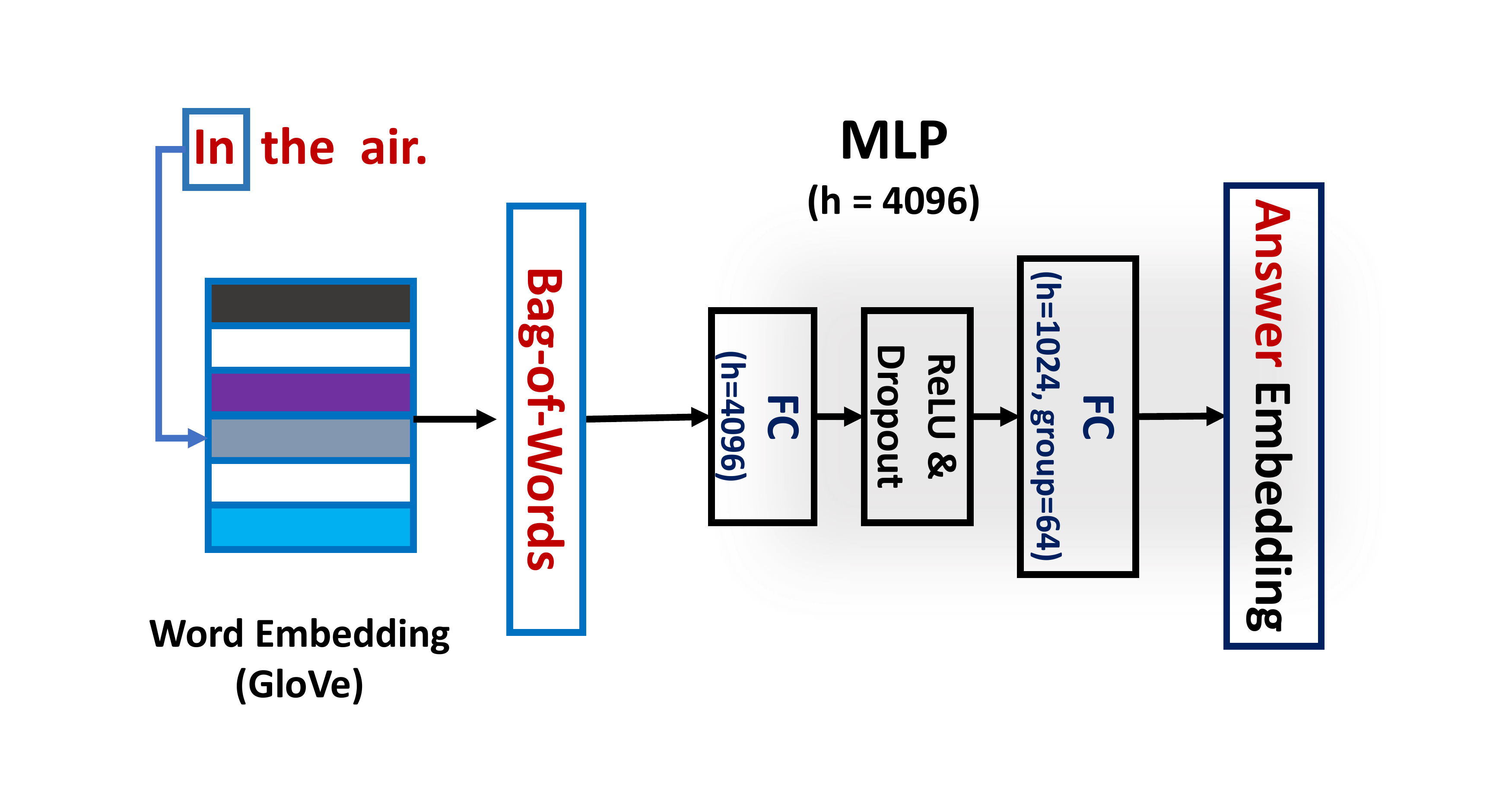}
	\caption{The multilayer perceptron (MLP) as $g_{\vphi}(a)$.  The average of  transformed word embeddings is first projected to the hidden space through a ReLU activation and then mapped to the embedding space. Dropout ($p$=0.5) is used for regularization. The same architecture is used for $f_{\vtheta}(i,q)$, except the input dimension is 2,348.}
	\vspace{-0.1in}
	\label{f_mlp}
\end{figure}

The architecture of the one-hidden-layer MLP for computing answer embedding is illustrated in Fig.~\ref{f_mlp}. The input is first mapped into the hidden space of 4,096 dimensions and then projected to a 1,024 dimensional embedding space. To reduce the number of parameters introduced in the MLP, we follow a similar practice suggested in \cite{xie2017aggregated} and apply a group-wise inner product to sparsify the weights. For both $f_{\vtheta}(i, q)$ and $g_{\vphi}(a)$, the output of MLP is scaled up by a factor 10. 

According to our ablation study in Section~\ref{s_semantics}, we set $\alpha$ (cf. eq.~(2) in the main text) to be multi-hot for VQA2, and use one-hot as $\alpha$ for all the other datasets. 

\begin{figure*}[tp]
	\centering
	\vspace{-0.1in}
	\includegraphics[width=0.975\textwidth]{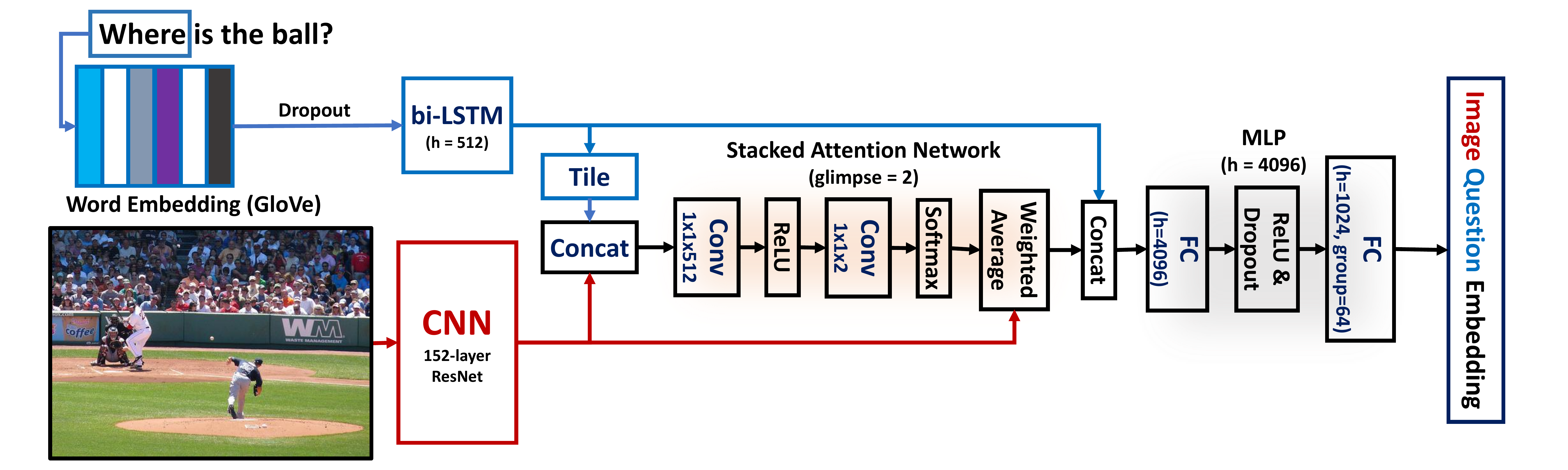}
	\caption{The stacked attention network (SAN) as $f_{\vtheta}(i, q)$. We follow the similar architecture as in~\cite{kazemi2017show} to obtain the visual semantic embedding of images and questions.}
	\vspace{-0.1in}
	\label{f_san}
\end{figure*}

\paragraph{SAN as $f_{\vtheta}(i, q)$}
Details about the stacked attention network (SAN) is shown in Fig.~\ref{f_san}. To represent a question, a single layer bidirectional LSTM (bi-LSTM) with the hidden dimension of 512 is used on top of the question GloVe word embeddings. Similarly to MLP setting, we fine-tune the question word embedding. At the same time, for image feature $i$, we extract the output of the last convolution layer from a 152-layer ResNet and obtain a feature tensor of dimensionality $14 \times 14 \times 2048$, as suggested in \cite{kazemi2017show}. A stacked attention module~\cite{yang2016stacked} with two glimpses is then used to obtain the question attended visual features, using both the outputs of question LSTM and ResNet-152 spatial visual feature. Next, a one-hidden-layer MLP (same architecture as previously mentioned) is used to embed the concatenated feature of questions and attended images into a 1,024-dimensional embedding space. Again, the output of the MLP is scaled up by a factor 10.

For our best performing model fPMC(SAN$\star$), we used the SAN as $f_{\vtheta}(i, q)$ and a two-layer bi-LSTM as answer embedding function $g_{\vphi}(a)$, with dimensionality of 512. For this bi-LSTM, we set the drop out rate to be 0.5 between the first and second LSTMs. We perform max fusion on the hidden states to obtain the holistic answer feature over the answer sentence. Both the output of $f_{\vtheta}(i, q)$ and $g_{\vphi}(a)$ are then scaled up by a factor of 10 and next used to produce the score through inner product for the $(i_n ,q_n, \calC_n)$ triplet. 

\paragraph{Configuration for competing methods}
For our classification model baseline (CLS), we use the same LSTM+SAN+MLP architecture as above, except that the output dimension is the total number of top-frequency answers. For the un-factorized PMC (uPMC), we concatenate the answer feature together with image and question features from SAN+LSTM and then input into a one-layer MLP with hidden dimensionality of 4096. It is then used to produce a singleton score for the input triplet. 

\paragraph{Optimization Details}
For all above methods, we train for 50 epochs on each dataset using Adam~\cite{kingma2014adam} optimization with initial learning rate of 0.001. We follows the same learning rate decay strategy suggested in~\cite{kazemi2017show}, which gives as follows:

\begin{equation}
	l_t = 0.5^{\frac{t}{t_{decay}}} \cdot l_0
\end{equation}

Here, $l_t$ denotes the learning rate at epoch $t$, $l_0$ is the initial learning rate. $t_{decay}$ represents the preset decay schedule, which is 15 in all our experiments. For fPMC we set the $A_o$ to be 3000 across all experiments; for uPMC, due to its large consumption of memory and computation inefficiency during training, we set the $A_o$ to be 300 for all settings (this is the largest feasible size of $A_o$ for uPMC(SAN) with reasonable computation and memory consumption). 

\begin{table}[b]
	\centering
	\caption{Detailed analysis of different $\alpha(a, d)$ for weighted likelihood. The reported number is the accuracy on VQA2 (validation).}
	\begin{tabular}{cc|c}\hline
		Method & Weighting Criterion & Acc. \\ \hline 
		\multirow{3}{*}{fPMC(SAN)} & {one-hot} & {58.0} \\
		& {multi-hot} & \textbf{60.0} \\
		& {WUPS} & 57.8  \\ \hline
	\end{tabular}
	\label{t_semantic}
\end{table}

\begin{table*}[t]
	\centering
	\caption{Analysis of cross dataset performance over Seen/Unseen answers using either CLS or PMC for Visual QA}
	\label{t_unseen}
	\begin{tabular}{c|ccc|ccc|ccc|ccc}\hline
		& \multicolumn{12}{|c}{Visual7W}  \\
		& \multicolumn{3}{|c}{CLS(SAN)} & \multicolumn{3}{c}{uPMC(SAN)} & \multicolumn{3}{c}{fPMC(SAN)}  & \multicolumn{3}{c}{fPMC(SAN$\star$)} \\
		& S & U & All & S & U & All & S & U & All & S & U & All \\ \hline
		{VQA2} \
		& 59.8 & 25.0 & 45.8 & 57.4 & 54.6 & 56.8 & 60.7 & 58.5 & {60.2} & 61.7 & 59.4 & 62.5 \\
		{qaVG} \
		& 63.4 & 25.0 & 58.9 & 66.7 & 45.3 & 66.0 & 69.1  & 47.7  & {68.4} & 70.2 & 46.9 & 69.5 \\
		\hline
	\end{tabular}
\end{table*}

\section{uPMC Vs. Triplet-based Methods}
\label{s_PMC_triplet}

We follow the exact multiple-choice (MC) setting of~\cite{chao2017being,jabri2016revisiting} to train MLP (with the $(i,q,a)$ triplet as input) on Visual7W. While getting good results on Visual7W (65.7\%), its transfer performance suffers (13.6\% to VQA2 and 30.2\% to qaVG). This is because in training, \cite{chao2017being,jabri2016revisiting} only differentiates between the correct answer and a few negative answers, not the entire universe of possible answers. Meanwhile, training the binary scoring function in~\cite{chao2017being,jabri2016revisiting} requires to carefully control the calibration between positive and negatives, which made it challenging when the number of negative answers scales up.

Therefore, we adapt their model to also utilize our PMC framework for training (i.e., uPMC(MLP)), which optimize stochastic multi-class cross-entropy with negative answers sampling. The transfer performance improves by a large margin. (Visual7W$\rightarrow$qaVG: improving from 30.2\% to 48.4\%.) 

\section{Semantic Knowledge in Weighted Likelihood}
\label{s_semantics}

As mentioned in section 4.4 of the main paper, we report in Table~\ref{t_semantic} the ablation study on using different weight function $\alpha(a, d)$ in the weighted likelihood formulation (cf. Eq. (2) of the main paper). We compare three different types of $\alpha(a, d)$ on VQA2:
\begin{itemize}
\item \textbf{one-hot}: Denote $t_n$ as the dominant answer in $\calC_n$. We set $\calC_n\leftarrow \{t_n\}$ (\ie, now $\calC_n$ becomes a singleton) and apply 
\begin{align}
\alpha(a, d) = \mathbb{I}[a=d] \text{ (cf. Eq. (3) of the main paper)}.  \nonumber
\end{align}
In this case, only one answer is considered positive to a $(i,q)$ pair. No extra semantic relationship is encoded.

\item \textbf{multi-hot}: We keep the given $\calC_n$ (the ten user annotations collected by VQA2; \ie $|\calC_n|=10$) and apply
\begin{align}
\alpha(a, d) = \mathbb{I}[a=d] \text{ (cf. Eq. (3) of the main paper)}  \nonumber
\end{align}
to obtain a multi-hot vector $\sum_{a\in\calC_n}\alpha(a, d)$ for soft weighting, leading to a loss similar to \cite{kazemi2017show,llievski2017simple}. 
 
\item \textbf{WUPS}: We again consider $\calC_n\leftarrow \{t_n\}$, but utilize the WUPS score~\cite{wu1994verbs, malinowski2014multi} (the range is [0, 1]) together with Eq. (6) of the main paper to define $\alpha(a, d)$. We set $\lambda=0.9$ and give $d$ which has $\textbf{WUPS}(a, d)=1$ a larger weight (\ie, 8).
\end{itemize} 

The results suggest that the multi-hot vector computed from multiple user annotations provides the best semantic knowledge among answers for learning the model. 

\section{Analysis with Seen/Unseen Answers}
\label{s_unseen}

Next, we present an analysis on transfer learning results, comparing the performance of methods over seen and unseen answer sets. Specifically, we study the transfer learning result from VQA2 and qaVG to Visual7W. Here, \textbf{seen (S)} refers to those multiple choices where at least one candidate answer is seen in the training vocabulary, and \textbf{unseen (U)} refers to those multiple choices where all the candidate answers are not observed in the training vocabulary. As shown in Table~\ref{t_unseen}, we see that our fPMC model performs better than the CLS model on both seen and unseen answer set. While CLS model obtains random performance (the random chance is 25 $\%$) on the unseen answer set, our fPMC model achieved  at least 20\% (in absolute value) better performance. In general. uPMC is also working well comparing to CLS. This performance improvement is gain mostly by taking answer semantics from the word vectors into account. 

\section{Visualization on Answer Embeddings}
\label{s_visualize}

\begin{figure*}[h]
	\begin{tabular}{c}
		\includegraphics[width=0.95\textwidth]{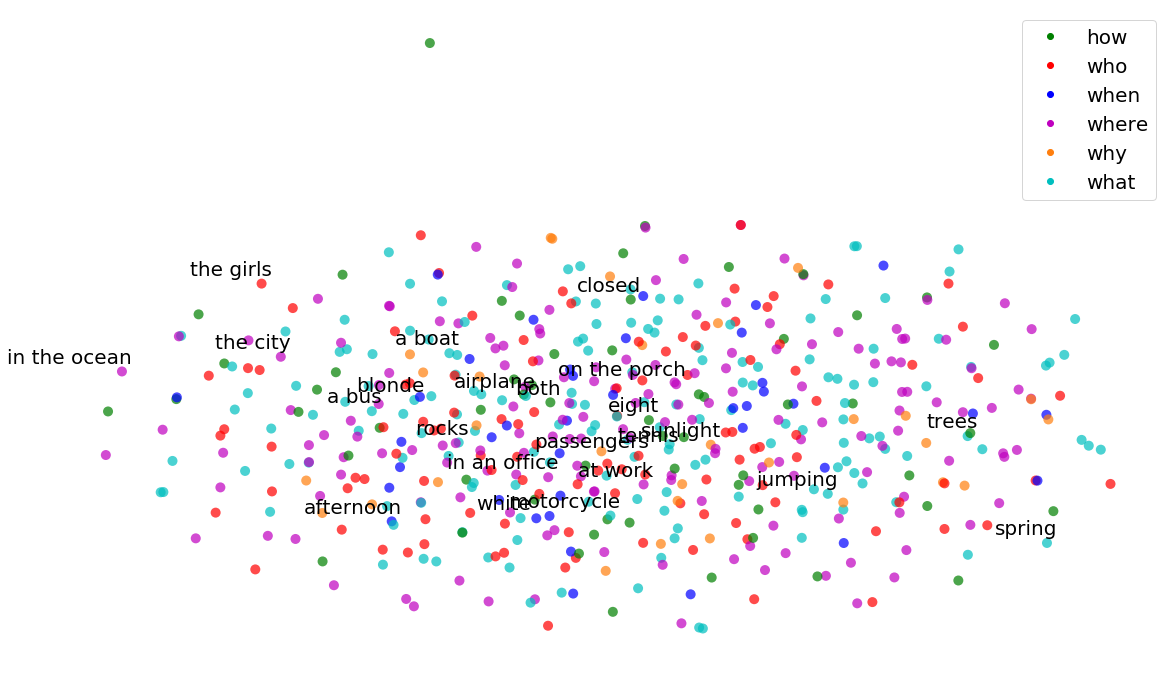} \\
		\textbf{ (a) Random initialized answer embedding } \\
	    \includegraphics[width=0.95\textwidth]{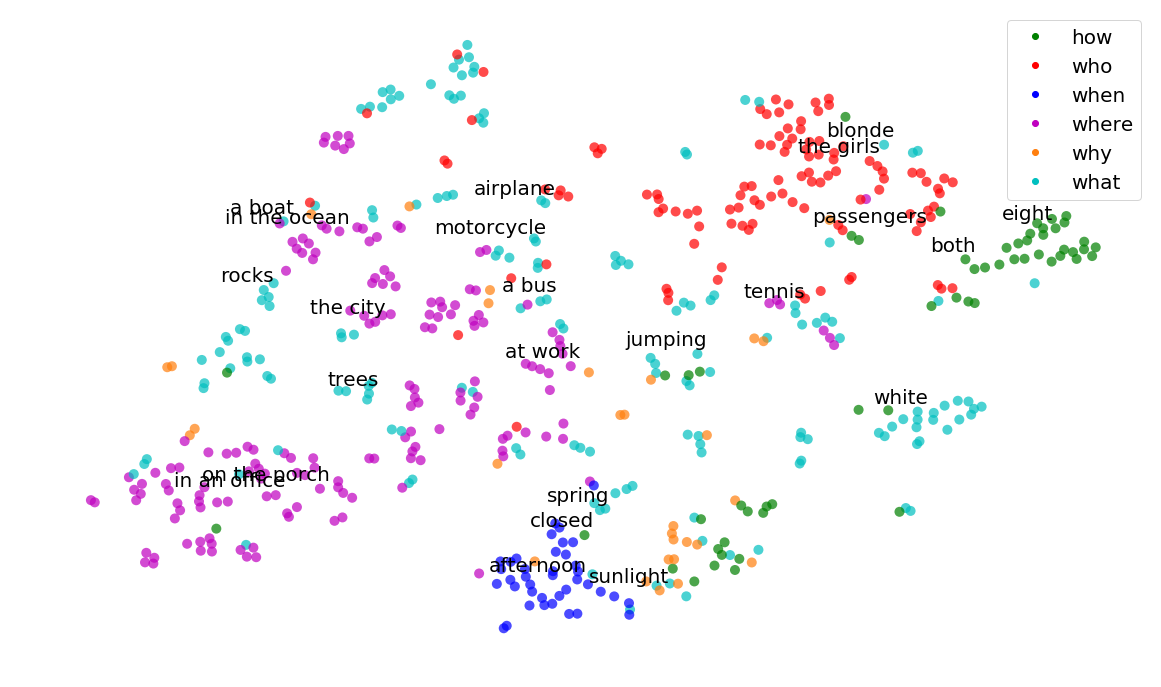} \\
	    \textbf{ (b) Learned answer embedding }
	\end{tabular}
	
	\caption{ \textbf{t-SNE visualization.} We randomly select 1000 answers from Visual7W and visualize  them in the initial answer embedding and learned answer embeddings. Each answer is marked with different colors according to their question types. (\eg when, how, who, where, why, what). To make the figure clear for reading, we randomly sub-sampled the text among those 1000 answers to visualize. }
	\label{f_tsne}
\end{figure*}

As promised in the main text, we provide the t-SNE visualization of the answer embedding. To better demonstrate the effectiveness of learning answer embedding, we re-train the answer embedding model with randomly initialized answer vectors. We provide visualization on both the initial answer embedding and learned answer embedding, to reflect the preservation of semantics and syntactics in the learned embedding. 

According to Fig.~\ref{f_tsne}, we can observe that a clear structure in the answer embedding are obtain in our learned embedding. While the random initialization of the embedding remains chaos, our learned embedding successfully provide both semantic and syntactic similarities between answers. For example, semantically similar answers such as \textbf{``airplane''} and \textbf{``motorcycle''} are close to each other, and syntactically similar answers like \textbf{``in an office''} and \textbf{``on the porch''} are close. Besides, we also observe that answers are clustered according to its majority question type, which meets our expectation for the answer embedding's structure. Here we take majority because one answer can be used for multiple questions of different types.

\section{Analysis on Answer Embeddings}
\label{s_baseline}

\begin{table}[h]
	\centering
	\small
	\tabcolsep 4pt
	\caption{Results for the baseline method that fix answer embedding as GloVe. (We show results with SAN as $f_{\vtheta}(i, q)$).}
	\label{t_baseline}
	\begin{tabular}{c|cc|cc|cc}\hline
		Target & \multicolumn{2}{|c}{VQA2} & \multicolumn{2}{|c}{Visual7W} & \multicolumn{2}{|c}{qaVG} \\
		Source & Fixed & Learning & Fixed & Learning & Fixed & Learning \\
		\hline 
		{VQA2}      & 57.5 & \textbf{60.0} & 47.5 & \textbf{60.2} & 37.6 & \textbf{54.8} \\
		\hline
	\end{tabular}
	\vspace{-5pt}
\end{table}

Finally, we provide results for an additional baseline algorithm where $f_{\vtheta}(i, q)$ directly maps to the fixed space of average GloVe answer representations. Here we need to keep the GloVe embedding fixed to enable transferability. Table~\ref{t_baseline} shows the results on the VQA2 dataset. We compare its performance to our approach of learning answer embedding with MLP as $g_{\vphi}(a)$ in terms of both in-domain and transfer learning performance---learning answer embeddings outperforms this simple baseline in all cases. Associated with the previous visualization results, we can conclude that learning answer embedding can effectively capture the semantic relationship between answers and image-question pairs while obtaining superior performance on both within-domain performance and transfer learning performance.

\end{document}